%% file: main.tex
\documentclass{article}

\usepackage{microtype}
\usepackage{graphicx}
\usepackage{subfigure}
\usepackage{booktabs} 
\usepackage{bbm} 
\usepackage{algcompatible}
\usepackage{algorithmic}
\usepackage{hyperref}

\usepackage[page]{appendix} 

\newcommand{\xxnote}[3]{}
\ifx\hidenotes\undefined
  \renewcommand{\xxnote}[3]{\color{#2}{#1: #3}}
\fi

\usepackage[accepted]{icml2020}

\input{math_commands.tex}

\icmltitlerunning{Generalized Hindsight for Reinforcement Learning}

\begin{document}

\twocolumn[
\icmltitle{Generalized Hindsight for Reinforcement Learning}



\icmlsetsymbol{equal}{*}

\begin{icmlauthorlist}
\icmlauthor{Alexander C. Li}{berkeley}
\icmlauthor{Lerrel Pinto}{berkeley,nyu}
\icmlauthor{Pieter Abbeel}{berkeley}
\end{icmlauthorlist}

\icmlaffiliation{berkeley}{Department of EECS, UC Berkeley, Berkeley, CA, USA}

\icmlaffiliation{nyu}{Computer Science, New York University, New York City, NY, USA}

\icmlcorrespondingauthor{Alexander C. Li}{alexli1@berkeley.edu}

\icmlkeywords{Reinforcement Learning, Multi-task Learning, Machine Learning, ICML}

\vskip 0.3in
]



\printAffiliationsAndNotice{}  

\begin{abstract}
One of the key reasons for the high sample complexity in reinforcement learning (RL) is the inability to transfer knowledge from one task to another. In standard multi-task RL settings, low-reward data collected while trying to solve one task provides little to no signal for solving that particular task and is hence effectively wasted. However, we argue that this data, which is uninformative for one task, is likely a rich source of information for other tasks. To leverage this insight and efficiently reuse data, we present Generalized Hindsight: an approximate inverse reinforcement learning technique for relabeling behaviors with the right tasks. Intuitively, given a behavior generated under one task, Generalized Hindsight returns a different task that the behavior is better suited for. Then, the behavior is relabeled with this new task before being used by an off-policy RL optimizer. Compared to standard relabeling techniques, Generalized Hindsight provides a substantially more efficient re-use of samples, which we empirically demonstrate on a suite of multi-task navigation and manipulation tasks. Videos and code can be accessed here: \href{https://sites.google.com/view/generalized-hindsight}{sites.google.com/view/generalized-hindsight}.

\end{abstract}

\input{introduction.tex}

\input{approach.tex}

\input{experiments.tex}

\input{related_work.tex}

\input{conclusion.tex}

\bibliography{main}
\bibliographystyle{icml2020}

\newpage
\onecolumn
\input{appendix.tex}





\end{document}

%% file: math_commands.tex
\usepackage{amsmath,amsfonts,bm}

\def\Secref#1{Section~\ref{#1}}

\def\eqref#1{equation~\ref{#1}}
\def\Eqref#1{Eq.~\ref{#1}}

\def\Algref#1{Algorithm~\ref{#1}}

\def\1{\bm{1}}

\DeclareMathAlphabet{\mathsfit}{\encodingdefault}{\sfdefault}{m}{sl}
\SetMathAlphabet{\mathsfit}{bold}{\encodingdefault}{\sfdefault}{bx}{n}

\newcommand{\E}{\mathbb{E}}

\DeclareMathOperator*{\argmax}{arg\,max}

\DeclareMathOperator*{\expect}{\E}

%% file: introduction.tex
\section{Introduction}

Model-free reinforcement learning (RL) combined with powerful function approximators has achieved remarkable success in games like Atari~\cite{mnih2015human} and Go~\cite{silver2017mastering}, and control tasks like walking~\cite{haarnoja2018soft} and flying~\cite{kaufmann2018deep}. However, a key limitation to these methods is their sample complexity. They often require millions of samples to learn simple locomotion skills, and sometimes even billions of samples to learn more complex game strategies. Creating general purpose agents will necessitate learning multiple such skills or strategies, which further exacerbates the inefficiency of these algorithms. On the other hand, humans (or biological agents) are not only able to learn a multitude of different skills, but from orders of magnitude fewer samples~\cite{karni1998acquisition}. So, how do we endow RL agents with this ability to learn efficiently across multiple tasks?

\begin{figure}[t!]
\centering
\includegraphics[width=\linewidth]{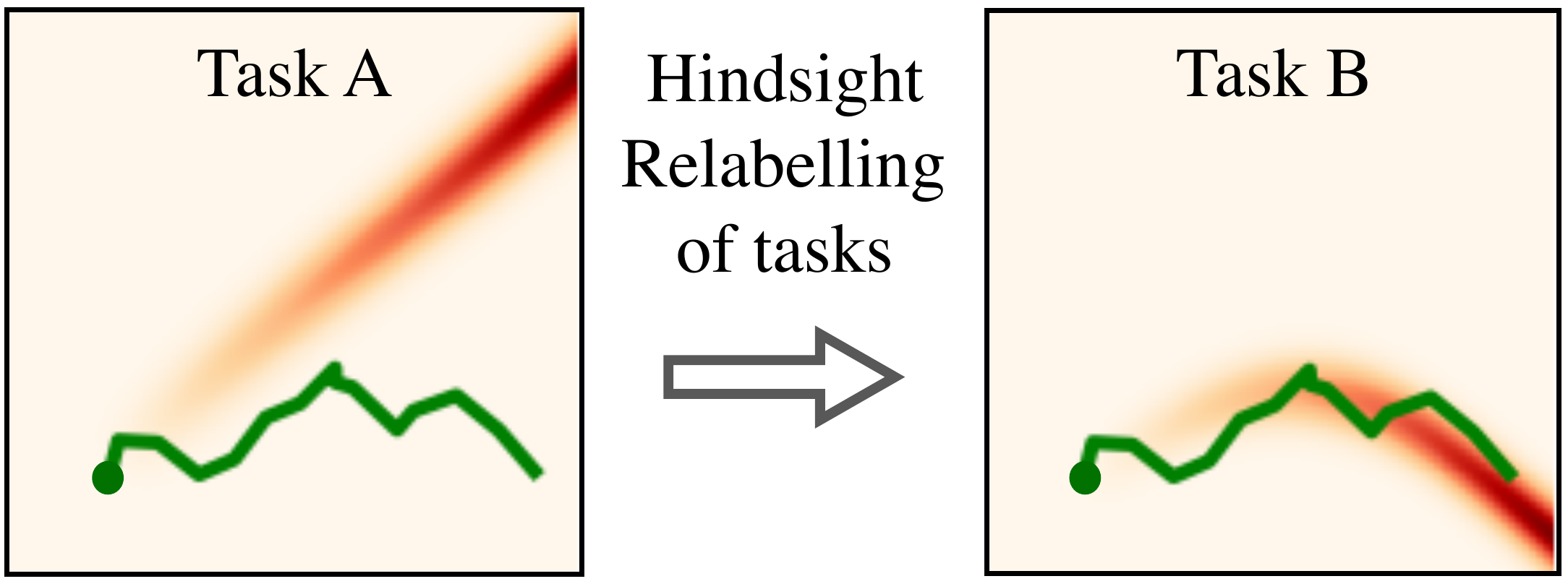}
\caption{A rollout can often provide very little information about how to perform a task A. In the trajectory-following task on the left, the trajectory (green) sees almost no reward signal (areas in red). However, in the multi-task setting where each target trajectory represents a different task, we can find another task B for which our trajectory is a ``pseudo-demonstration." This hindsight relabeling provides high reward signal and enables sample-efficient learning.}
\label{fig:illustration}
\end{figure}

One key hallmark of biological learning is the ability to learn from mistakes. In RL, mistakes made while solving a task are only used to guide the learning of that particular task. But data seen while making these mistakes often contain a lot more information. In fact, extracting and re-using this information lies at the heart of most efficient RL algorithms. Model-based RL re-uses this information to learn a dynamics model of the environment. However for several domains, learning a robust model is often more difficult than directly learning the policy~\cite{duan2016benchmarking}, and addressing this challenge continues to remain an active area of research~\cite{nagabandi2018neural}. Another way to re-use low-reward data is off-policy RL, where in contrast to on-policy RL, data collected from an older policy is re-used while optimizing the new policy. But in the context of multi-task learning, this is still inefficient (\Secref{section:experiments}) since data generated from one task cannot effectively inform a different task. Towards solving this problem, recent work~\cite{andrychowicz2017hindsight} focus on extracting even more information through \textit{hindsight}.

In goal-conditioned settings, where tasks are defined by a sparse goal, HER~\cite{andrychowicz2017hindsight} relabels the desired goal, for which a trajectory was generated, to a state seen in that trajectory. Therefore, if the goal-conditioned policy erroneously reaches an incorrect goal instead of the desired goal, we can re-use this data to teach it how to reach this incorrect goal. Hence, a low-reward trajectory under one desired goal is converted to a high-reward trajectory for the unintended goal. This new relabelling provides a strong supervision and produces significantly faster learning. However, a key assumption made in this framework is that goals are a sparse set of states that need to be reached. This allows for efficient relabeling by simply setting the relabeled goals to the states visited by the policy. But for several real world problems like energy-efficient transport, or robotic trajectory tracking, rewards are often complex combinations of desirables rather than sparse objectives. So how do we use hindsight for general families of reward functions?

\begin{figure*}[t!]
\centering
\includegraphics[width=0.9\textwidth]{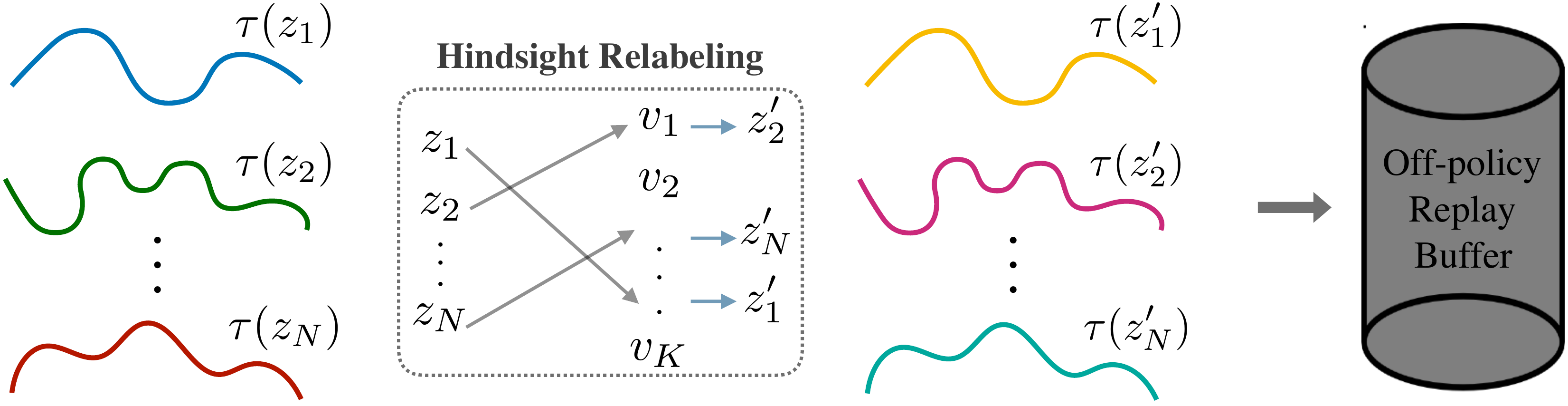}
\caption{Trajectories $\tau(z_i)$, collected trying to maximize $r(\cdot|z_i)$, may contain very little reward signal about how to solve their original tasks. Generalized Hindsight checks against randomly sampled ``candidate tasks" $\{v_i\}_{i=1}^K$ to find different tasks $z'_i$ for which these trajectories are ``pseudo-demonstrations." Using off-policy RL, we can obtain more reward signal from these relabeled trajectories.}
\label{fig:intro}
\end{figure*}

In this paper, we build on the ideas of goal-conditioned hindsight and propose \textbf{Generalized Hindsight}. Here, instead of performing hindsight on a task-family of sparse goals, we perform hindsight on a task-family of reward functions. Since dense reward functions can capture a richer task specification, GH allows for better re-utilization of data. Note that this is done along with solving the task distribution induced by the family of reward functions. However for relabeling, instead of simply setting visited states as goals, we now need to compute the reward functions that best explain the generated data. To do this, we draw connections from Inverse Reinforcement Learning (IRL), and propose an approximate IRL relabeling algorithm we call AIR. Concretely, AIR takes a new trajectory and compares it to $K$ randomly sampled tasks from our distribution. It selects the task for which the trajectory is a ``pseudo-demonstration," i.e. the trajectory achieves higher performance on that task than any of our previous trajectories. This ``pseudo-demonstration" can then be used to quickly learn how to perform that new task. We go into detail on good selection algorithms in \Secref{section:experiments}, and show an illustrative example of the relabeling process in \autoref{fig:illustration}. We test our algorithm on several multi-task control tasks, and find that AIR consistently achieves higher asymptotic performance using as few as 20\% of the environment interactions as our baselines. We also introduce a computationally more efficient version that also achieves higher asymptotic performance than our baselines.

In summary, we present three key contributions in this paper: (a) we extend the ideas of hindsight to the generalized reward family setting; (b) we propose AIR, a relabeling algorithm using insights from IRL; and (c) we demonstrate significant improvements in multi-task RL on a suite of multi-task navigation and manipulation tasks.

%% file: approach.tex
\section{Background}
Before discussing our method, we briefly introduce some background and formalism for the RL algorithms used. A more comprehensive introduction to RL can be found in \citet{kaelbling1996reinforcement} and \citet{sutton1998reinforcement}.

\subsection{Reinforcement Learning}
In this paper we deal with continuous space Markov Decision Processes $\mathcal{M}$ that can be represented as the tuple $\mathcal{M}\equiv(\mathcal{S},\mathcal{A},\mathcal{P},r,\gamma, \mathbb{S})$, where $\mathcal{S}$ is a set of continuous states and $\mathcal{A}$ is a set of continuous actions, $\mathcal{P}: \mathcal{S} \times \mathcal{A} \times \mathcal{S} \rightarrow \mathbb{R}$ is the transition probability function, $r: \mathcal{S} \times \mathcal{A} \rightarrow \mathbb{R}$ is the reward function, $\gamma$ is the discount factor, and $\mathbb{S}$ is the initial state distribution. 

An episode for the agent begins with sampling $s_0$ from the initial state distribution $\mathbb{S}$. 
At every timestep $t$, the agent takes an action $a_t=\pi(s_t)$ according to a policy $\pi:\mathcal{S} \rightarrow \mathcal{A}$. 
At every timestep $t$, the agent gets a reward $r_t=r(s_t,a_t)$, and the state transitions to $s_{t+1}$, which is sampled according to probabilities $\mathcal{P}(s_{t+1} | s_t,a_t)$. 
The goal of the agent is to maximize the expected return $\expect_{\mathbb{S}}[R_0|\mathbb{S}]$, where the return is the discounted sum of the future rewards $R_t=\sum^{\infty}_{i=t}\gamma^{i-t}r_i$. The $Q$-function is defined as $Q^{\pi}(s_t,a_t)=E[R_t|s_t,a_t]$. In the partial observability case, the agent takes actions based on the partial observation, $a_t=\pi(o_t)$, where $o_t$ is the observation corresponding to the full state $s_t$.

\subsection{Off Policy RL using Soft Actor Critic}

Generalized Hindsight requires an off-policy RL algorithm to perform relabeling. One popular off-policy algorithm for learning deterministic continuous action policies is Deep Deterministic Policy Gradients (DDPG)~\cite{lillicrap2015continuous}. The algorithm maintains two neural networks: the policy (also called the actor) $\pi_\theta:\mathcal{S} \rightarrow \mathcal{A}$ (with neural network parameters $\theta$) and a $Q$-function approximator (also called the critic) $Q_\phi^{\pi}:\mathcal{S} \times \mathcal{A} \rightarrow \mathbb{R}$ (with neural network parameters $\phi$). 

During training, episodes are generated using a noisy version of the policy (called behaviour policy), 
e.g. $\pi_b(s) = \pi(s) + \mathcal{N}(0,1)$, where $\mathcal{N}$ is the Normal distribution noise. 
The transition tuples $(s_t,a_t,r_t,s_{t+1})$ encountered during training are stored in a replay buffer~\cite{mnih2015human}. Training examples sampled from the replay buffer are used to optimize the critic. By minimizing the Bellman error loss $\mathcal{L}_c=(Q(s_t,a_t)-y_t)^2$, where $y_t=r_t + \gamma Q(s_{t+1},\pi(s_{t+1}))$, the critic is optimized to approximate the $Q$-function. The actor is optimized by minimizing the loss $\mathcal{L}_a=-\expect_s[Q(s,\pi(s))]$. The gradient of $\mathcal{L}_a$ with respect to the actor parameters is called the deterministic policy gradient ~\cite{silver2014deterministic} and can be computed by backpropagating through the combined critic and actor networks. 
To stabilize the training, the targets for the actor and the critic $y_t$ are computed on separate versions of the actor and critic networks, 
which change at a slower rate than the main networks. A common practice is to use a Polyak averaged~\cite{polyak1992acceleration} version of the main network. Soft Actor Critic (SAC) \cite{haarnoja2018soft} builds on DDPG by adding an entropy maximization term in the reward. Since this encourages exploration and empirically performs better than most actor-critic algorithms, we use SAC for our experiments, although Generalized Hindsight is compatible with any off-policy RL algorithm.

\subsection{Multi-Task RL}
The goal in multi-task RL is to not just solve a single MDP $\mathcal{M}$, but to solve to solve a distribution of MDPs $\mathcal{M}(z)$, where $z$ is the task-specification drawn from the task distribution $z\sim\mathcal{T}$. Although $z$ can  parameterize different aspects of the MDP, we are specially interested in the different reward functions. Hence, our distribution of MDPs is now $\mathcal{M}(z)\equiv(\mathcal{S},\mathcal{A},\mathcal{P},r(\cdot|z),\gamma, \mathbb{S})$. Thus, a different $z$ implies a different reward function under the same dynamics $\mathcal{P}$ and start state $z$. One may view this representation as a generalization of the goal-conditioned RL setting~\cite{schaul2015universal}, where the reward family is restricted to $r(s,a|z=g)=-d(s,z=g)$. Here $d$ represents the distance between the current state $s$ and the desired goal $g$. In sparse goal-conditioned RL, where hindsight has previously been applied~\cite{andrychowicz2017hindsight}, the reward family is further restricted to $r(s,a|z=g)=[d(s,z=g)<\epsilon]$. Here the agent gets a positive reward only when $s$ is within $\epsilon$ of the desired goal $g$.

\subsection{Hindsight Experience Replay (HER)}

HER~\cite{andrychowicz2017hindsight} is a simple method of manipulating the replay buffer used in off-policy RL algorithms that allows it to learn state-reaching policies more efficiently with sparse rewards. After experiencing some episode $s_0,s_1,...,s_T$, every transition $s_t\rightarrow s_{t+1}$ along with the goal for this episode is usually stored in the replay buffer. However with HER, the experienced transitions are also stored in the replay buffer with different goals. These additional goals are states that were achieved later in the episode. Since the goal being pursued does not influence the environment dynamics, one can replay each trajectory using arbitrary goals, assuming we use an off-policy RL algorithm to optimize~\cite{precup2001off}.

\subsection{Inverse Reinforcement Learning (IRL)}
In IRL~\cite{ng2000algorithms}, given an expert policy $\pi_E$ or more practically, access to demonstrations $\tau_E$ from $\pi_E$, we want to recover the underlying reward function $r^*$ that best explains the expert behaviour. Although there are several methods that tackle this problem~\cite{ratliff2006maximum,abbeel2004apprenticeship,ziebart2008maximum}, the basic principle is to find $r^*$ such that:
\begin{equation}
    \E[\sum_{t=0}^{T-1} \gamma r^*(s_t)|\pi_E] \geq \E[\sum_{t=0}^{T-1} \gamma r^*(s_t)|\pi] \;\; \forall \pi
\end{equation}
We use the framework of IRL to guide our \textit{Approximate IRL} relabeling strategy for Generalized Hindsight.

\section{Generalized Hindsight}
\subsection{Overview}
Given a multi-task RL setup, i.e. a distribution of reward functions $r(.|z)$, our goal is to maximize the expected reward across the task distribution $z\sim\mathcal{T}$ through optimizing our policy $\pi$:
\begin{equation}
    \expect_{z\sim\mathcal{T}}[R(\pi|z)]
\end{equation}
Here, $R(\pi|z)=\sum^{T-1}_{t=0}\gamma^{t}r(s_t,a_t\sim\pi(s_t|z)|z)$ represents the cumulative discounted reward under the reward parameterization $z$ and the conditional policy $\pi(.|z)$. One approach to solving this problem would be the straightforward application of RL to train the $z-$ conditional policy using the rewards from $r(.|z)$. However, this fails to re-use the data collected under one task parameter $z$ $(s_t,a_t)\sim\pi(.|z)$ to a different parameter $z'$. In order to better use and share this data, we propose to use hindsight relabeling, which is detailed in Algorithm \ref{alg:hindsight}.

The core idea of hindsight relabeling is to convert the data generated from the policy under one task $z$ to a different task. Given the relabeled task $z'=\texttt{relabel}(\tau(\pi(.|z)))$, where $\tau$ represents the trajectory induced by the policy $\pi(.|z)$, the state transition tuple $(s_t,a_t,r_t(.|z),s_{t+1})$ is converted to the relabeled tuple $(s_t,a_t,r_t(.|z'),s_{t+1})$. This relabeled tuple is then added to the replay buffer of an off-policy RL algorithm and trained as if the data generated from $z$ was generated from $z'$. If relabeling is done efficiently, it will allow for data that is sub-optimal under one reward specification $z$, to be used for the better relabeled specification $z'$. In the context of sparse goal-conditioned RL, where $z$ corresponds to a goal $g$ that needs to be achieved, HER~\cite{andrychowicz2017hindsight} relabels the goal to states seen in the trajectory, i.e. $g'\sim\tau(\pi(.|z=g))$. This labeling strategy, however, only works in sparse goal conditioned tasks. In the following section, we describe two relabeling strategies that allow for a generalized application of hindsight.
%

\begin{algorithm}
   \caption{$\texttt{Generalized Hindsight}$}
   \label{alg:hindsight}
\begin{algorithmic}[1]
    \STATE {\bfseries Input:} Off-policy RL algorithm $\mathbb{A}$, strategy $\mathbb{S}$ for choosing suitable task variables to relabel with, reward function $r: \mathcal{S} \times \mathcal{A} \times \mathcal{Z} \rightarrow \mathbb{R}$
    \FOR{episode $=1$ to $M$}
        \STATE Sample a task variable $z$ and an initial state $s_0$
        \STATE Roll out policy on $z$ for $T$ steps, yielding trajectory $\tau$
        \STATE Find set of new tasks to relabel with: $Z := \mathbb{S}(\tau)$
        \STATE Store original transitions in replay buffer: 
        \Statex $\quad$ $(s_t, a_t, r(s_t, a_t, z), s_{t+1}, z)$
        \FOR{$z' \in Z$}
                \STATE Store relabeled transitions in replay buffer:
                \Statex $\quad \quad$ $(s_t, a_t, r(s_t, a_t, z'), s_{t+1}, z')$ 
        \ENDFOR
        \STATE Perform $n$ steps of policy optimization with $\mathbb{A}$
    \ENDFOR
\end{algorithmic}
\end{algorithm}

\subsection{Approximate IRL Relabeling (AIR)}

\begin{algorithm}
   \caption{$\texttt{$\mathbb{S}_{IRL}$: Approximate IRL}$}
   \label{alg:approxirl}
\begin{algorithmic}[1]
    \STATE {\bfseries Input:} Trajectory $\tau = (s_0, a_0, ..., s_T)$, cached trajectories $\mathcal{D} = \{(s_0, a_0, ..., s_T)\}_{i=1}^N$, reward function $r: \mathcal{S} \times \mathcal{A} \times \mathcal{Z} \rightarrow \mathbb{R}$, number of candidate task variables to try: $K$, number of task variables to return: $m$
    \STATE Sample set of candidate tasks $Z = \{v_j\}_{j=1}^K$, where $v_j \sim \mathcal{T}$
    \Statex \textbf{Approximate IRL Strategy:}
    \FOR{$v_j \in Z$}
        \STATE \textbf{Calculate trajectory reward} for $\tau$ and the trajectories in $\mathcal{D}$: $R(\tau|v_j) := \sum_{t=0}^T \gamma^t r(s_t, a_t, v_j)$ 
        \STATE \textbf{Calculate percentile estimate:} 
        \Statex $\quad \quad  \hat{P}(\tau, v_j) = \frac{1}{n} \sum_{i=1}^N \mathbbm{1}\{R(\tau|v_j) \geq R(\tau_i|v_j)\}$
    \ENDFOR
    \STATE \textbf{return} $m$ tasks $v_j$ with highest percentiles $\hat{P}(\tau, v_j)$
\end{algorithmic}
\end{algorithm}

The goal of computing the optimal reward parameter, given a trajectory is closely tied to the Inverse Reinforcement Learning (IRL) setting. In IRL, given demonstrations from an expert, we can retrieve the reward function the expert was optimized for. At the heart of these IRL algorithms, a reward specification parameter $z'$ is optimized such that
\begin{equation}
\label{eq:irl_orig}
    R(\tau_E|z')\geq R(\tau'|z') \; \forall \, \tau'
\end{equation} 
where $\tau_E$ is an expert trajectory. Inspired by the IRL framework, we propose the \textit{Approximate IRL} relabeling seen in \Algref{alg:approxirl}. We can use a buffer of past trajectories to find the task $z'$ on which our current trajectory does better than the older ones. Intuitively this can be seen as an approximation of the right hand side of ~\Eqref{eq:irl_orig}. Concretely, we want to relabel a new trajectory $\tau$, and have $N$ previously sampled trajectories along with $K$ randomly sampled candidate tasks $v_k$. Then, the relabeled task for trajectory $\tau$ is computed as:
\begin{equation}
    z' = \argmax_{k} \frac{1}{N} \sum_{j=1}^N \mathbbm{1}\{R(\tau|v_k) \geq R(\tau_j|v_k)\}
\end{equation}
The relabeled $z'$ for $\tau$ maximizes its percentile among the $N$ most recent trajectories collected with our policy. One can also see this as an approximation of max-margin IRL~\cite{ratliff2006maximum}. 
One potential challenge with large $K$ is that many $v_k$ will have the same percentile. To choose between these potential task relabelings, we add tiebreaking based on the advantage estimate 
\begin{equation}
    \hat{A}(\tau, z) = R(\tau|z) - V^\pi(s_0, z)
\end{equation}
Among candidate tasks $v_k$ with the same percentile, we take the tasks that have higher advantage estimate. From here on, we will refer to Generalized Hindsight with Approximate IRL Relabeling as AIR. 

\subsection{Advantage Relabeling}

\begin{figure*}[t!]
\centering
\subfigure[PointTrajectory]{\label{fig:pointmass_env}\includegraphics[width=0.19\linewidth]{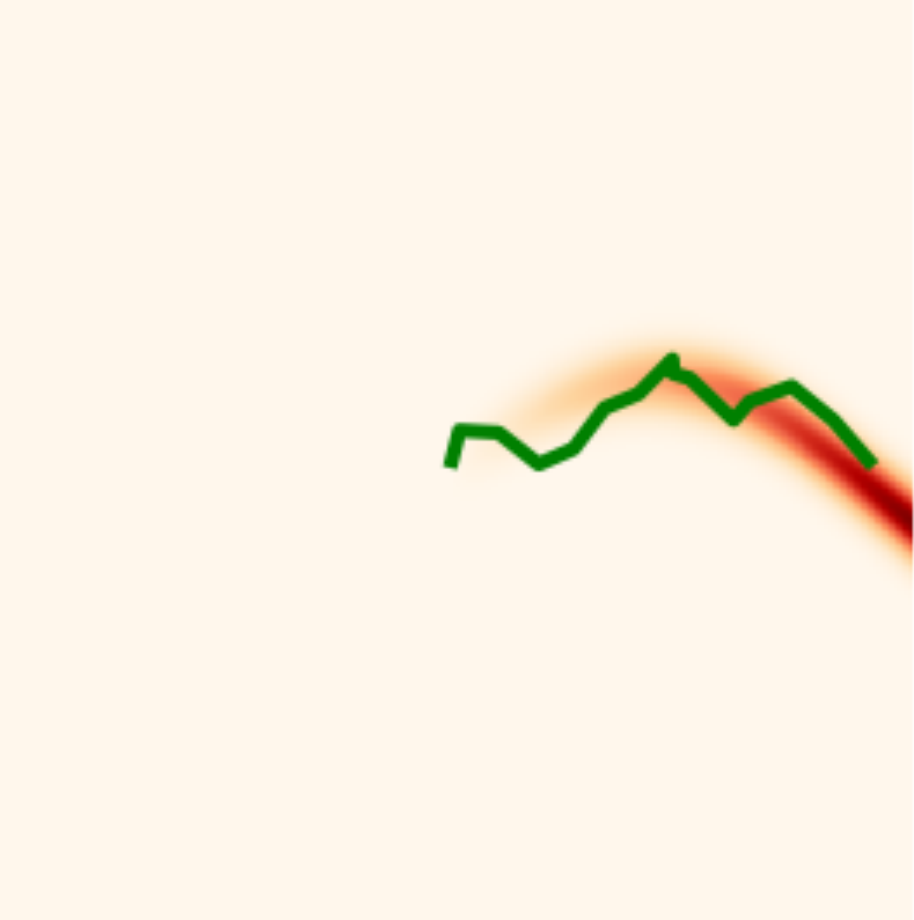}}
\subfigure[PointReacher]{\label{fig:pointgoal_env}\includegraphics[width=0.19\linewidth]{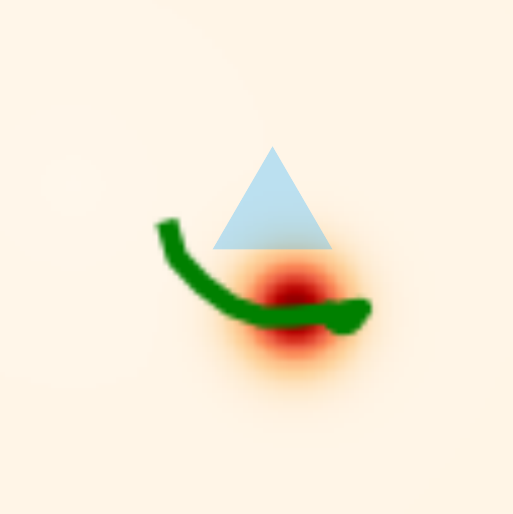}}
\subfigure[Fetch]{\includegraphics[width=0.19\linewidth]{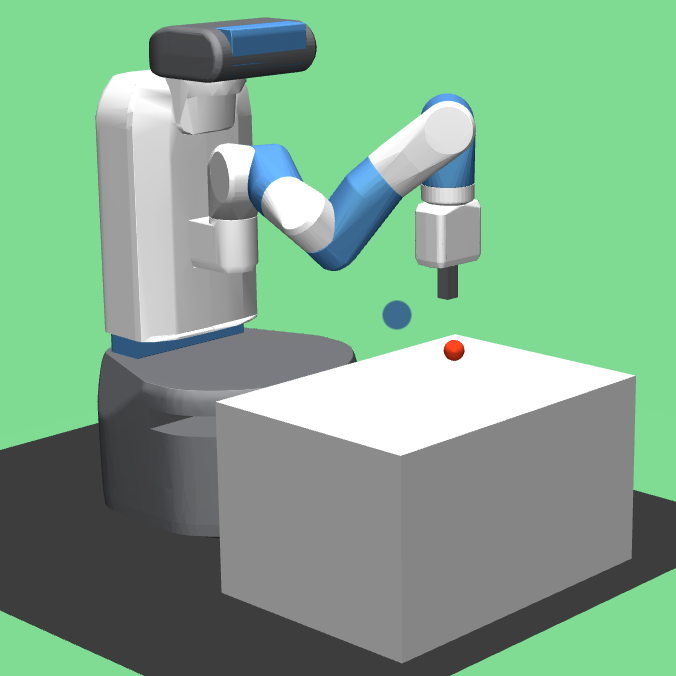}}
\subfigure[HalfCheetahMultiObj]{\includegraphics[width=0.19\linewidth]{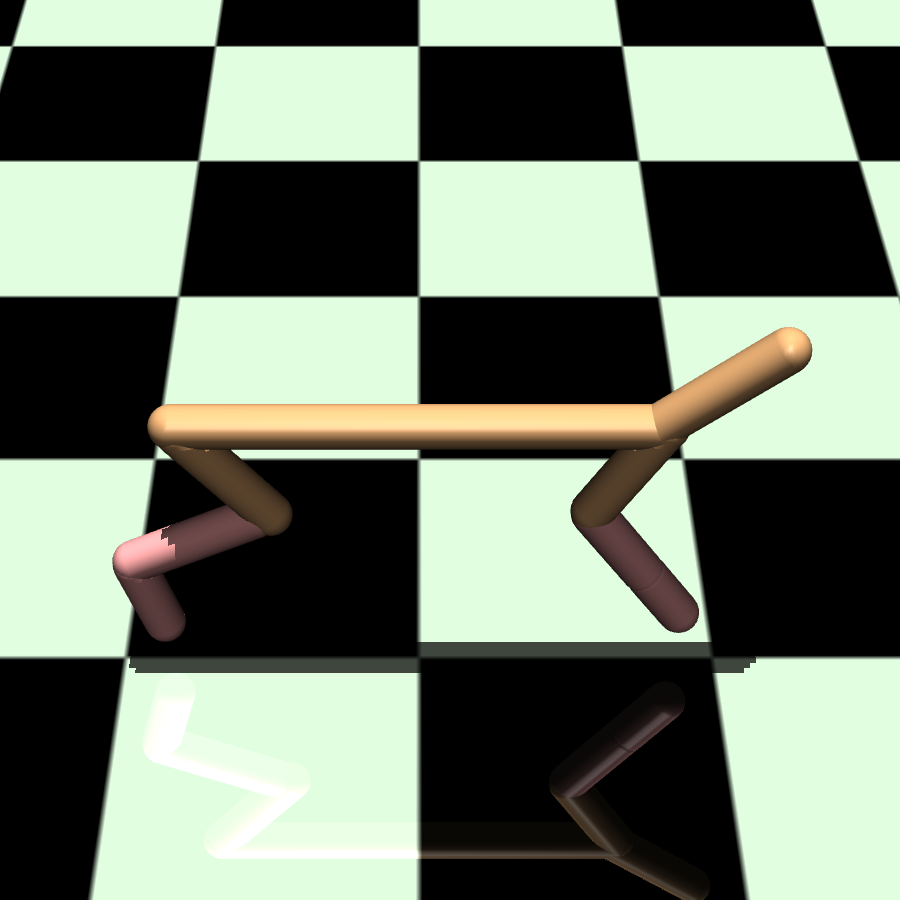}}
\subfigure[AntDirection]{\includegraphics[width=0.19\linewidth]{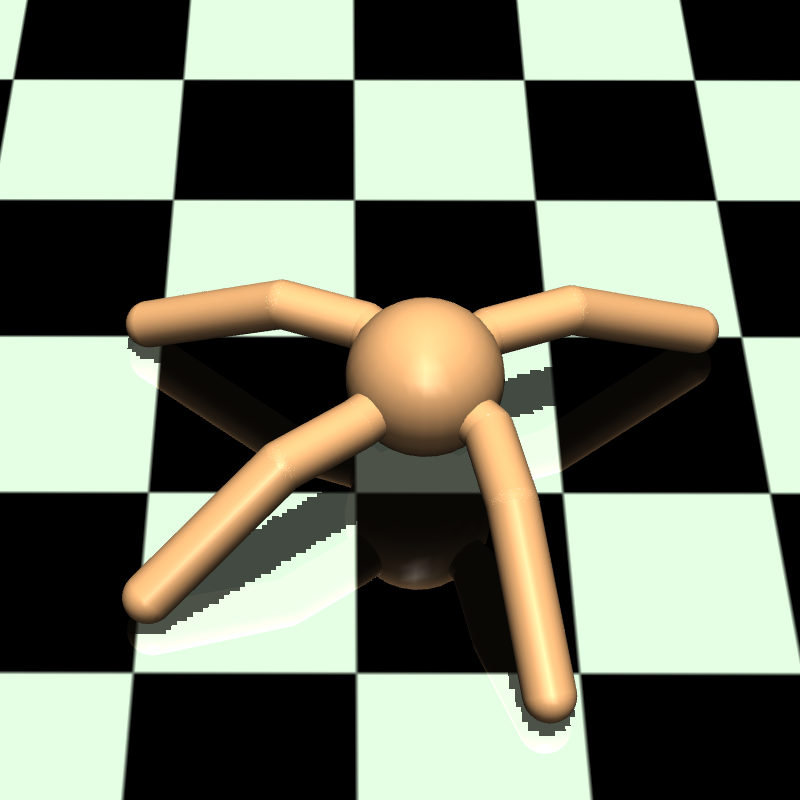}}
\caption{Environments we report comparisons on. PointTrajectory requires a 2D pointmass to follow a target trajectory; PointReacher requires moving the pointmass to a goal location, while avoiding an obstacle and modulating its energy usage. In (b), the red circle indicates the goal location, while the blue triangle indicates an imagined obstacle to avoid. Fetch has the same reward formulation as PointReacher, but requires controlling the noisy Fetch robot in 3 dimensions. HalfCheetah requires learning running in both directions, flipping, jumping, and moving efficiently. AntDirection requires moving in a target direction as fast as possible.}
\label{fig:learningcurves}
\end{figure*}

One potential problem with AIR is that it requires $O(NT)$ time to compute the relabeled task variable for each new trajectory, where $N$ is the number of past trajectories compared to, and $T$ is the horizon. A relaxed version of AIR could significantly reduce computation time, while maintaining relatively high-accuracy relabeling. One way to do this is to use the \textit{Maximum-Reward} relabeling objective. Instead of choosing from our $K$ candidate tasks $v_k\sim\mathcal{T}$ by selecting for high percentile (\autoref{eq:irl_orig}), we could relabel based on the cumulative trajectory reward:
\begin{equation}
    z'=\argmax_{v_k} \{R(\tau|v_k)\}
\end{equation}
However, one challenge with simply taking the \textit{Maximum-Reward} relabel is that different reward parameterizations may have different scales which will bias the relabels to a specific $z$. Say for instance there exists a task in the reward family $v_j$ such that $r(.|v_j) = 1 + \max_{i\neq j} r(.|v_i)$. Then, $v_j$ will always be the relabeled reward parameter irrespective of the trajectory $\tau$. Hence, we should not only care about the $v_k$ that maximizes reward, but select $v_k$ such that $\tau$'s likelihood under the trajectory distribution drawn from the optimal $\pi^*(.|v_k)$ is high. To do this, we can simply select $z'$ based on the advantage term that we used to tiebreak for AIR. 
\begin{equation}
    z'_i = \argmax_{k} R(\tau|v_k) - V^\pi(s_0, v_k)
\end{equation}
We call this \textit{Advantage} relabeling (Algorithm \ref{alg:advantage}), a more efficient, albeit less accurate, version of AIR. Empirically, \textit{Advantage} relabeling often performs as well as AIR, but requires the value function $V^\pi$ to be more accurate than it has to be in AIR. We reuse the twin $Q$-networks from SAC as our value function.
\begin{equation}
    V^\pi(s, z) = \min(Q_1(s, \pi(s|z), z), Q_2(s, \pi(s|z), z))
\end{equation}

\begin{algorithm}
   \caption{$\texttt{$\mathbb{S}_A$: Trajectory Advantage}$}
   \label{alg:advantage}
\begin{algorithmic}[1]
    \STATE Repeat steps 1 \& 2 from Algorithm \ref{alg:approxirl}
    \Statex \textbf{Advantage Relabeling Strategy:}
    \FOR{$v_j \in Z$}
        \STATE \textbf{Calculate trajectory reward:} 
        \Statex $\quad \quad R(\tau|v_j) := \sum_{t=0}^T \gamma^t r(s_t, a_t, v_j)$
        \STATE \textbf{Calculate advantage estimate of the trajectory}: 
        \Statex $\quad \quad \hat{A}(\tau, v_j) = R(\tau|v_j) - V^\pi(s_0, v_j)$
    \ENDFOR
    \STATE \textbf{return} $m$ tasks $z_j$ with highest advantages $\hat{A}(\tau, z_j)$
\end{algorithmic}
\end{algorithm}

%% file: experiments.tex
\section{Experimental Evaluation}

\label{section:experiments}

In this section, we describe our environment settings along with a discussion of our central hypothesis: Does relabeling improve performance? 

\begin{figure*}[t!]
\centering
\includegraphics[width=\linewidth]{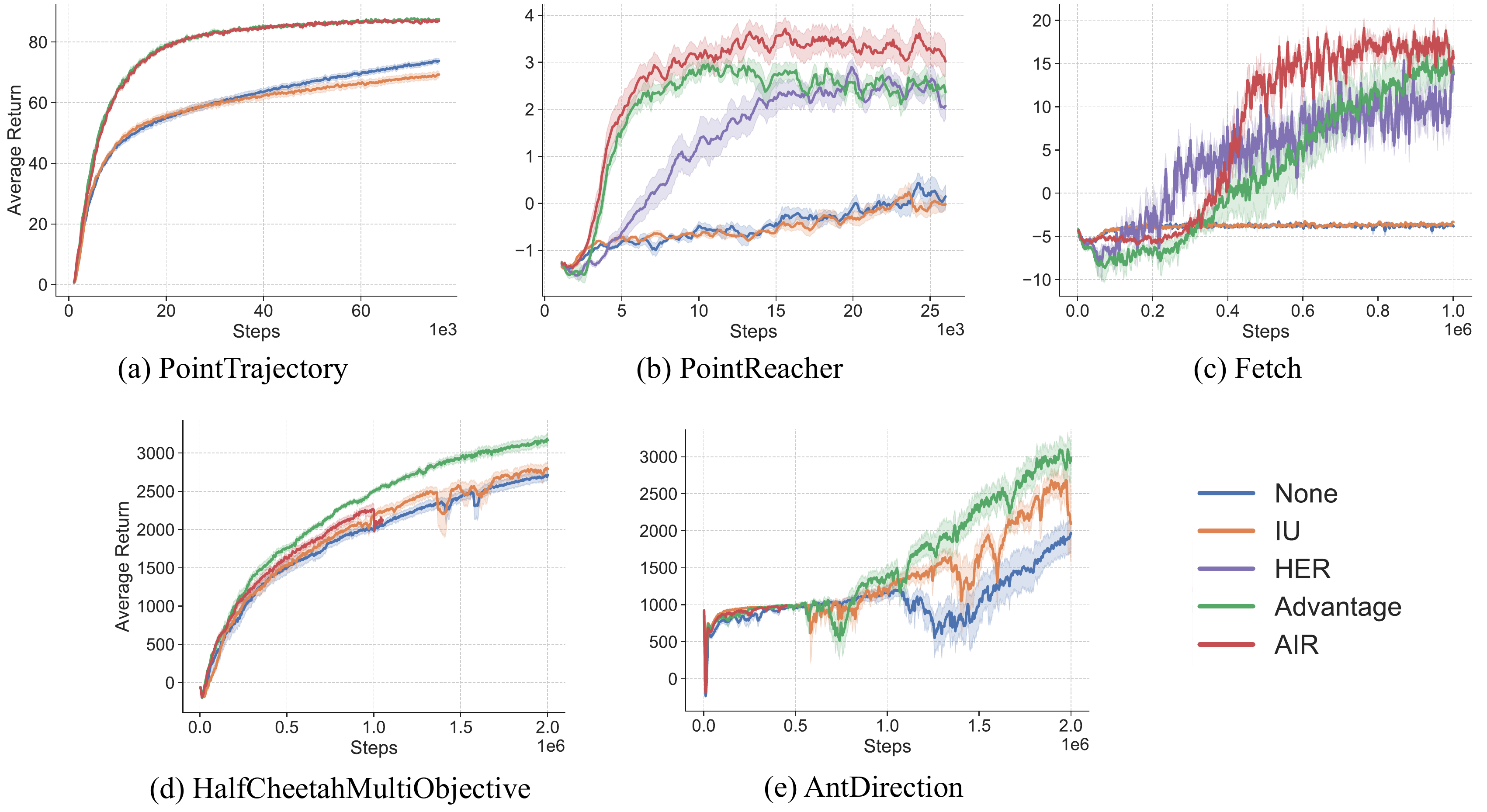}
\caption{Learning curves comparing Generalized Hindsight algorithms to baseline methods.  For environments with a goal-reaching component, we also compare to HER. In (a), AIR learning curve obscures the Advantage learning curve. In (d) and (e), where we use $N$ = 500 for AIR, AIR takes much longer to run than the other methods. 10 seeds were used for all runs.}
\label{fig:learningcurves}
\end{figure*}

\subsection{Environment setting}
Multi-task RL with a generalized family of reward parameterizations does not have existing benchmark environments. However, since sparse goal-conditioned RL has benchmark environments~\cite{plappert2018multi}, we build on their robotic manipulation framework to make our environments. The key difference in the environment setting between ours and \citet{plappert2018multi} is that in addition to goal reaching, we have a dense reward parameterization for practical aspects of manipulation like energy consumption~\cite{meike2011energy} and safety~\cite{calinon2010learning}. These environments will be released for open-source access. The five environments we use are as follows:

\begin{enumerate}
  \item \textbf{PointTrajectory}: 2D pointmass with $(x,y)$ observations and $(dx, dy)$ position control for actions. The goal is to follow a target trajectory parameterized by $z \in \mathcal{Z} \subseteq \mathbb{R}^3$. Figure \autoref{fig:pointmass_env} depicts an example trajectory in green, overlaid on the reward heatmap defined by some specific task $z$. 
  \item \textbf{PointReacher}: 2D pointmass with $(x,y)$ observations and $(dx, dy)$ position control for actions. This environment has high reward around the goal position $(x_g, y_g)$ and low reward around an obstacle location $(x_{obst}, y_{obst})$. The 6-dimensional task vector is $z = (x_g, y_g, x_{obst}, y_{obst}, u, v)$, where $u$ and $v$ control the weighting between the goal rewards, obstacle rewards, and action magnitude penalty. 
  \item \textbf{Fetch}: Here we adapt the Fetch environment from OpenAI Gym~\cite{brockman2016openai}, with $(x,y,z)$ end-effector position as observations and noisy position control for actions. We use the same parameterized reward function as in PointReacher that includes energy and safety specifications. 
  \item \textbf{HalfCheetahMultiObjective}: HalfCheetah-V2 from OpenAI Gym, with 17-dimensional observations and 6-dimensional actions for torque control. The task variable $z = (w_{vel}, w_{rot}, w_{height}, w_{energy}) \in \mathcal{Z} = S^3$ controls the weights on the forward velocity, rotation speed, height, and energy rewards.
  \item \textbf{AntDirection}: Ant-V2 from OpenAI gym, with 111-dimensional observations and 8-dimensional actions for torque control. The task variable $z \in [-180^\circ, +180^\circ]$ parameterizes the target direction. The reward function is $r(\cdot|z) = ||\text{velocity}||_2 \times \mathbbm{1}\{\text{velocity angle within 15 degrees of z}\}$.
\end{enumerate}

\subsection{Does Relabeling Help?}
To understand the effects of relabeling, we compare our technique with the following standard baseline methods:
\begin{itemize}
    \item No relabeling (None): as done in \cite{yu2019meta}, we train with standard SAC without any relabeling step.
    \item Intentional-Unintentional Agent (IU) \cite{cabi2017intentional}: when there is only a finite number of tasks, IU relabels a trajectory with every task variable. Since our space of tasks is continuous, we relabel with random $z' \sim \mathcal{T}$. This allows for information to be shared across tasks, albeit in a more diluted form. 
    \item HER: for goal-conditioned tasks, we use HER to relabel the goal portion of the latent with the future relabeling strategy. 
\end{itemize}
We compare the learning performance for AIR and Advantage Relabeling with these baselines on our suite of environments in \autoref{fig:learningcurves}. On all tasks, AIR and Advantage Relabeling outperform the baselines in both sample-efficiency and asymptotic performance. Both of our relabeling strategies outperform the Intentional-Unintentional Agent, implying that selectively relabeling trajectories with a few carefully chosen $z'$ is more effective than relabeling with many random tasks. Collectively, these results show that AIR can greatly improve learning performance, even on highly dense environments such as HalfCheetahMultiObjective, where learning signal is readily available. 

\begin{figure}[!t]
\centering
\subfigure[Stationary]{\includegraphics[width=0.23\linewidth]{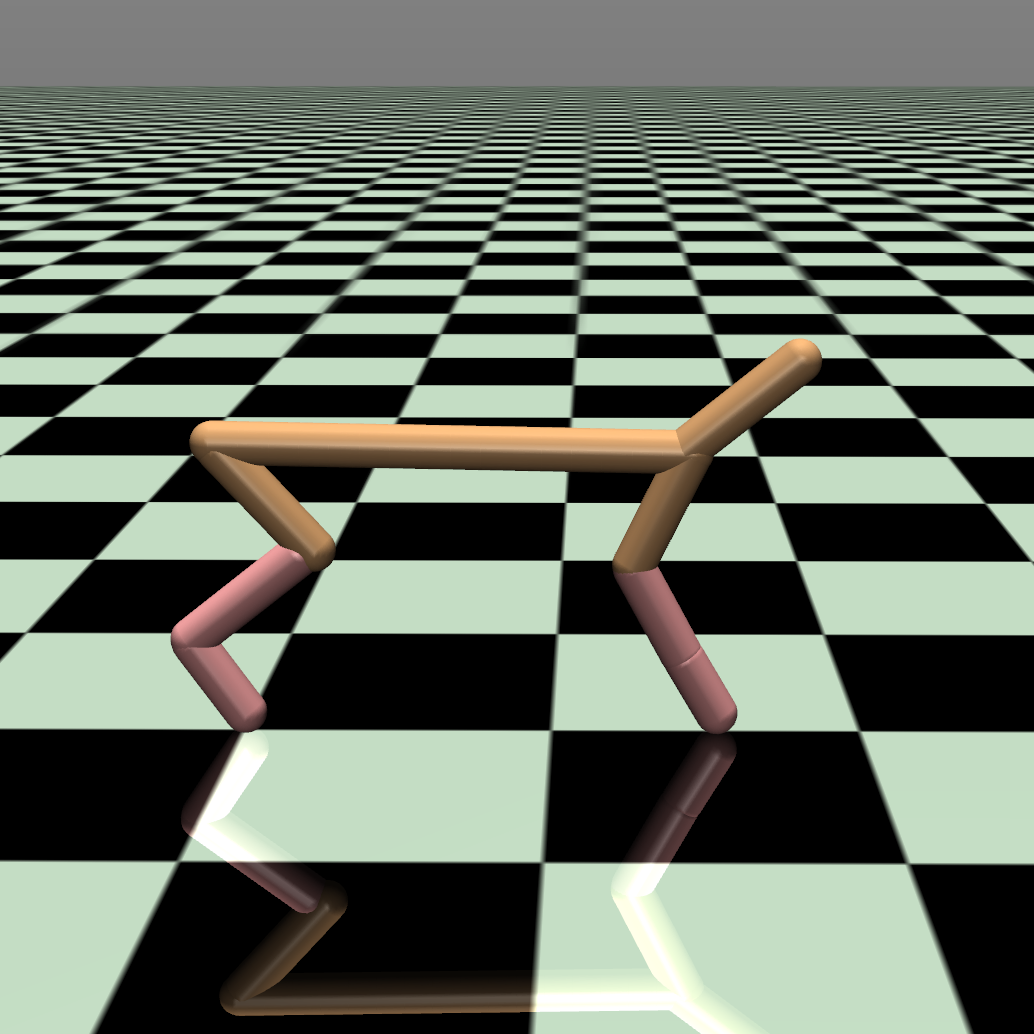}}
\subfigure[Forward]{\includegraphics[width=0.23\linewidth]{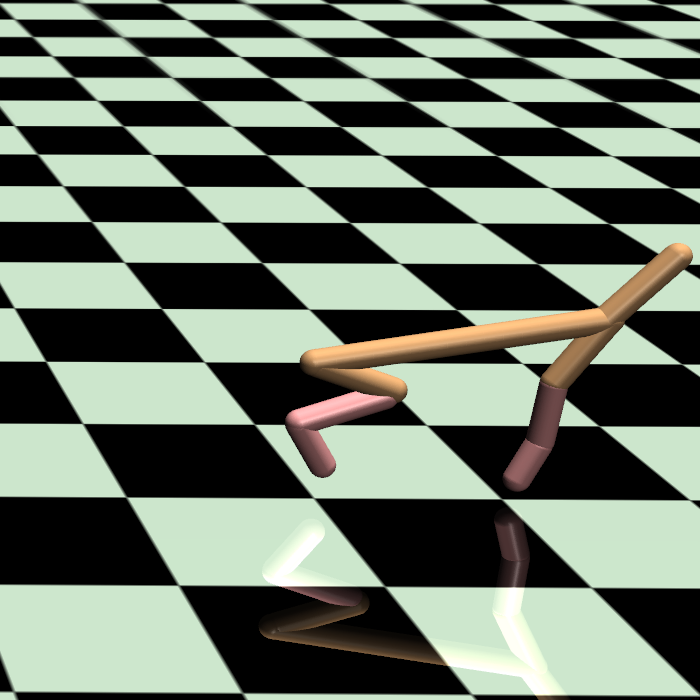}}
\subfigure[Back]{\includegraphics[width=0.23\linewidth]{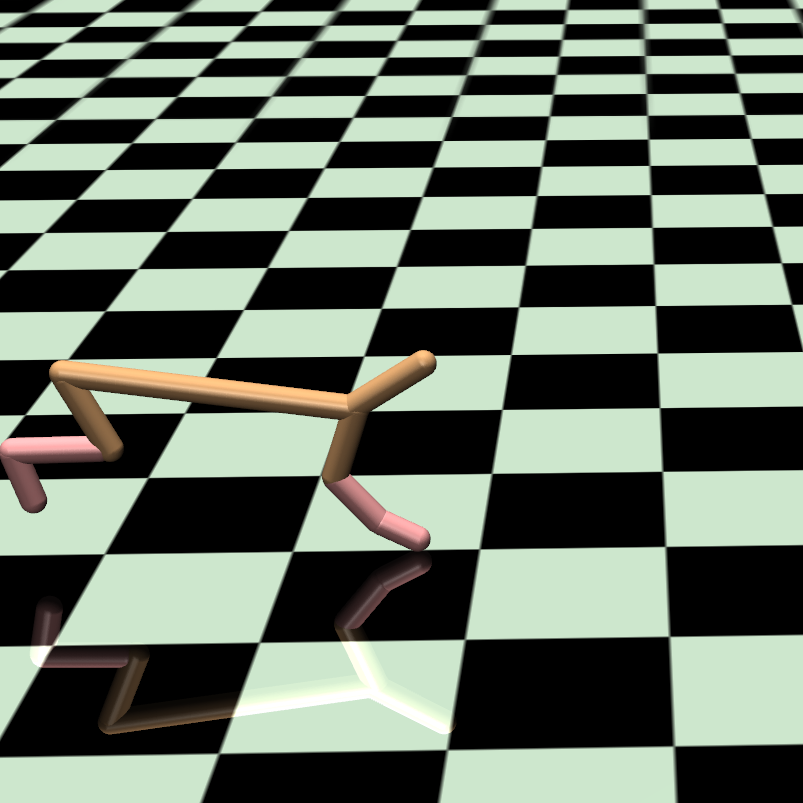}}
\subfigure[Frontflip]{\includegraphics[width=0.23\linewidth]{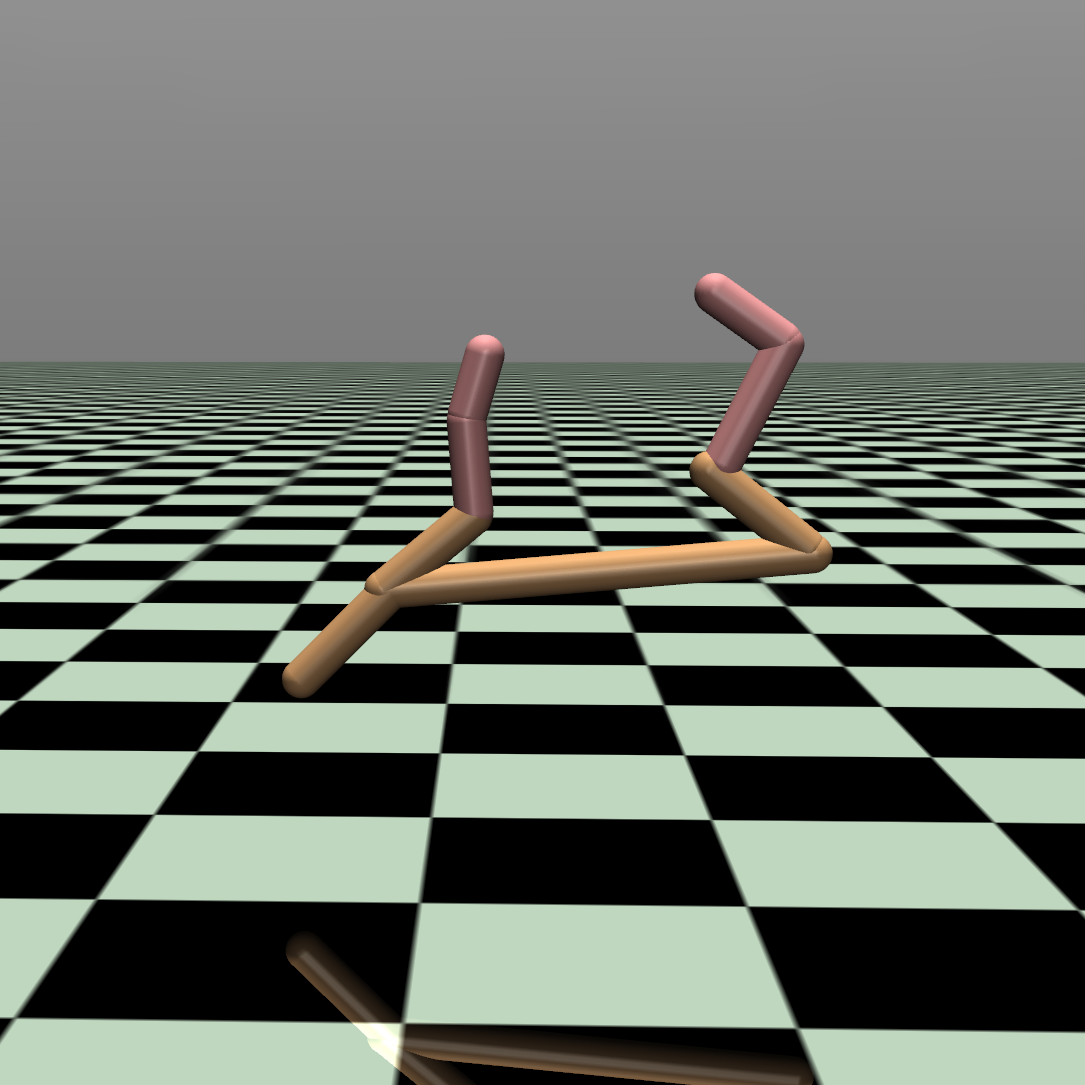}}
\subfigure[Left]{\includegraphics[width=0.23\linewidth]{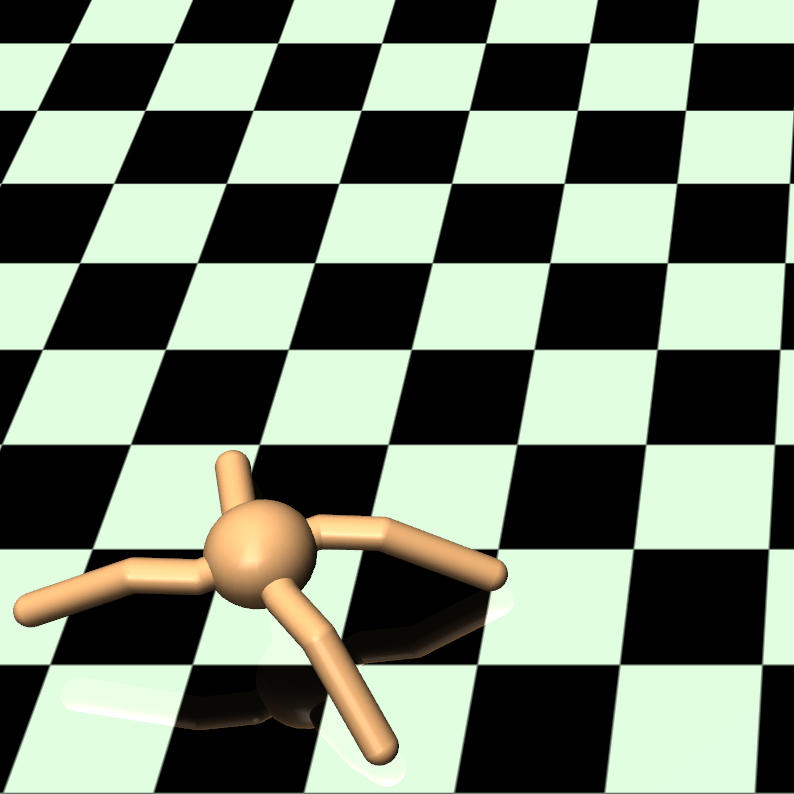}}
\subfigure[Right]{\includegraphics[width=0.23\linewidth]{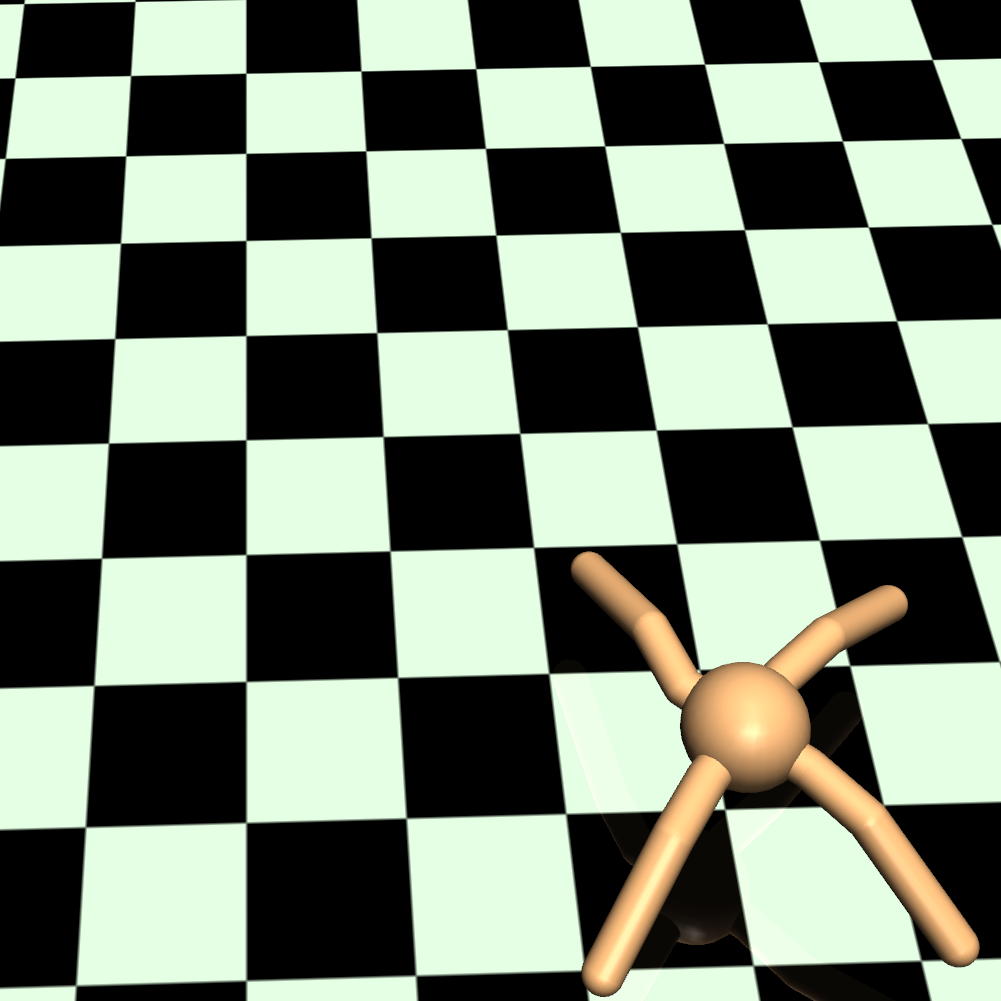}}
\subfigure[Top Right]{\includegraphics[width=0.23\linewidth]{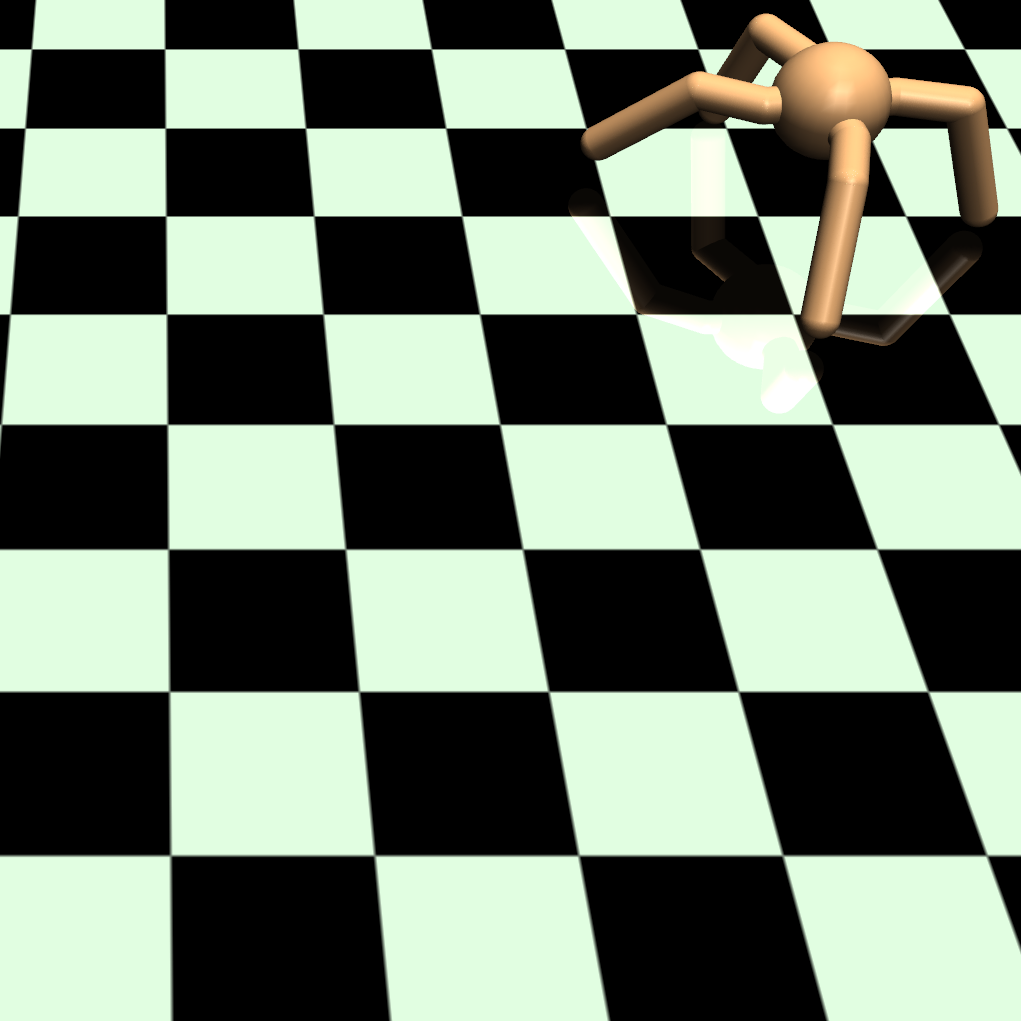}}
\subfigure[Top Left]{\includegraphics[width=0.23\linewidth]{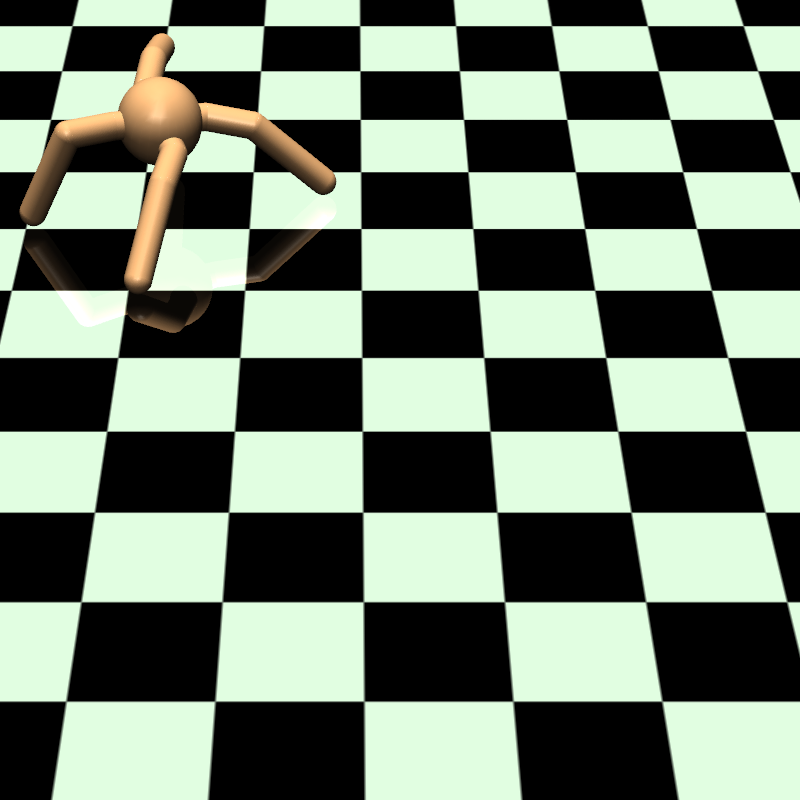}}
\caption{The agents efficiently learn a wide range of behaviors. On HalfCheetahMultiObjective, the robot can stay still to conserve energy, run quickly forwards and backwards, and do a frontflip. On AntDirection, the robot can run quickly in any given direction.}
\label{fig:behavior}
\end{figure}

\subsection{How does generalized relabeling compare to HER?}
\begin{figure}[!t]
\centering
\includegraphics[width=\linewidth]{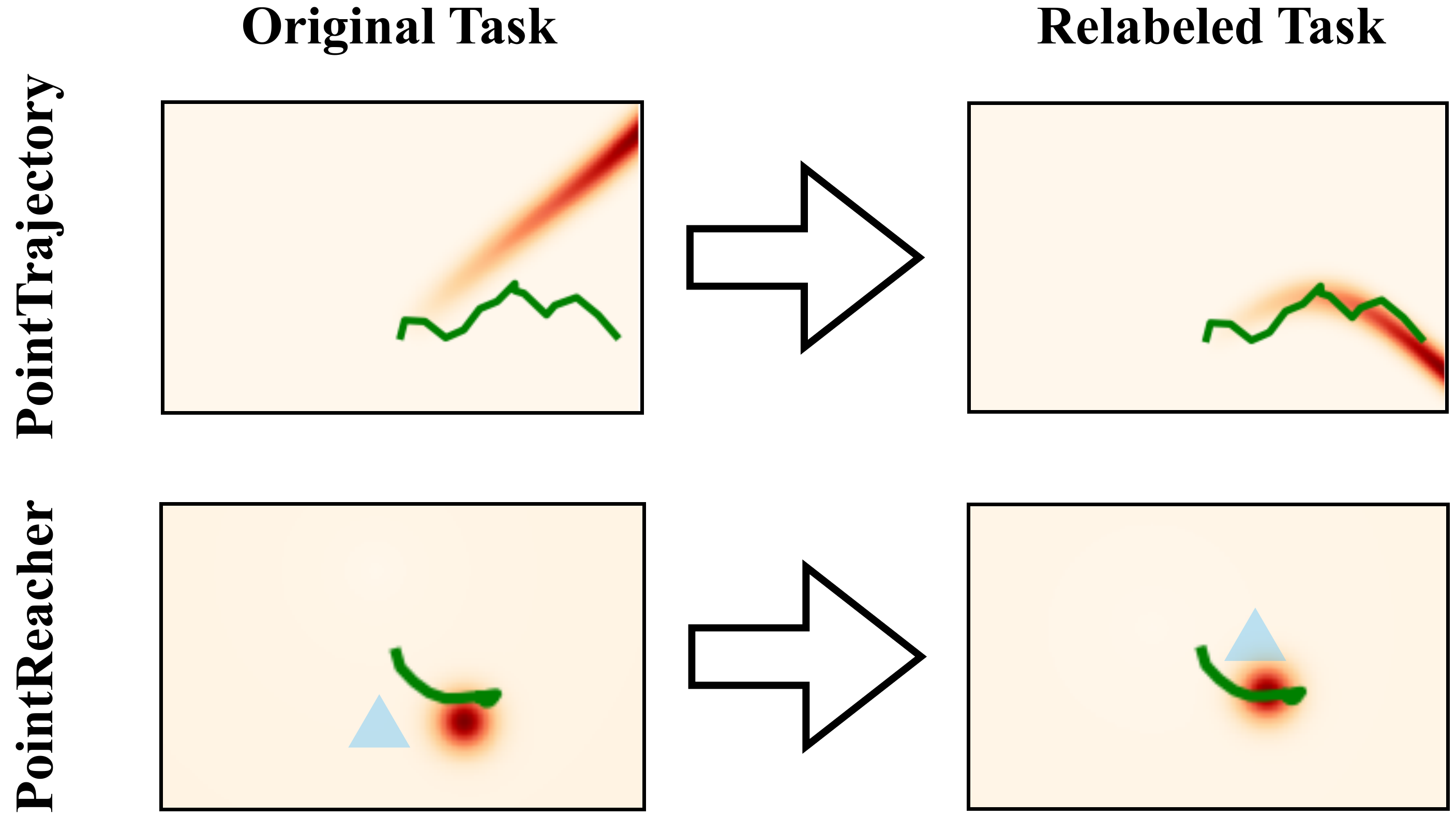}
\caption{Red denotes areas of high reward, for following a target trajectory (top) or reaching a goal (bottom). Blue indicates areas of negative reward, where an obstacle may be placed. On both environments, relabeling finds tasks on which our trajectory has high reward signal. 
On PointReacher, AIR does not place the obstacle arbitrarily far. It places the relabeled obstacle within the curve of the trajectory, since this is the only way that the curved path would be better than a straight-line path (that would come close to the relabeled obstacle). }
\label{fig:relabelingexample}
\end{figure}

HER is, by design, limited to goal-reaching environments. For environments such as HalfCheetahMultiObjective, HER cannot be applied to relabel the weights on velocity, rotation, height, and energy. However, we can compare AIR with HER on the partially goal-reaching environments PointReacher and Fetch. \autoref{fig:learningcurves} shows that AIR achieves higher asymptotic performance than HER on both these environments. \autoref{fig:relabelingexample} demonstrates on PointReacher how AIR can better choose the non-goal-conditioned parts of the task. Both HER and AIR understand to place the relabeled goal around the terminus of the trajectory. However, only AIR understands that the imagined obstacle should be placed above the goal, since this trajectory becomes an optimal example of how to reach the new goal while avoiding the obstacle. HER (as well as the Intentional-Unintentional Agent) offer no such specificity, either leaving the obstacle in place or randomly placing it. 

\begin{figure*}[!t]
\centering
\subfigure[Approximate IRL]{\includegraphics[width=0.33\linewidth]{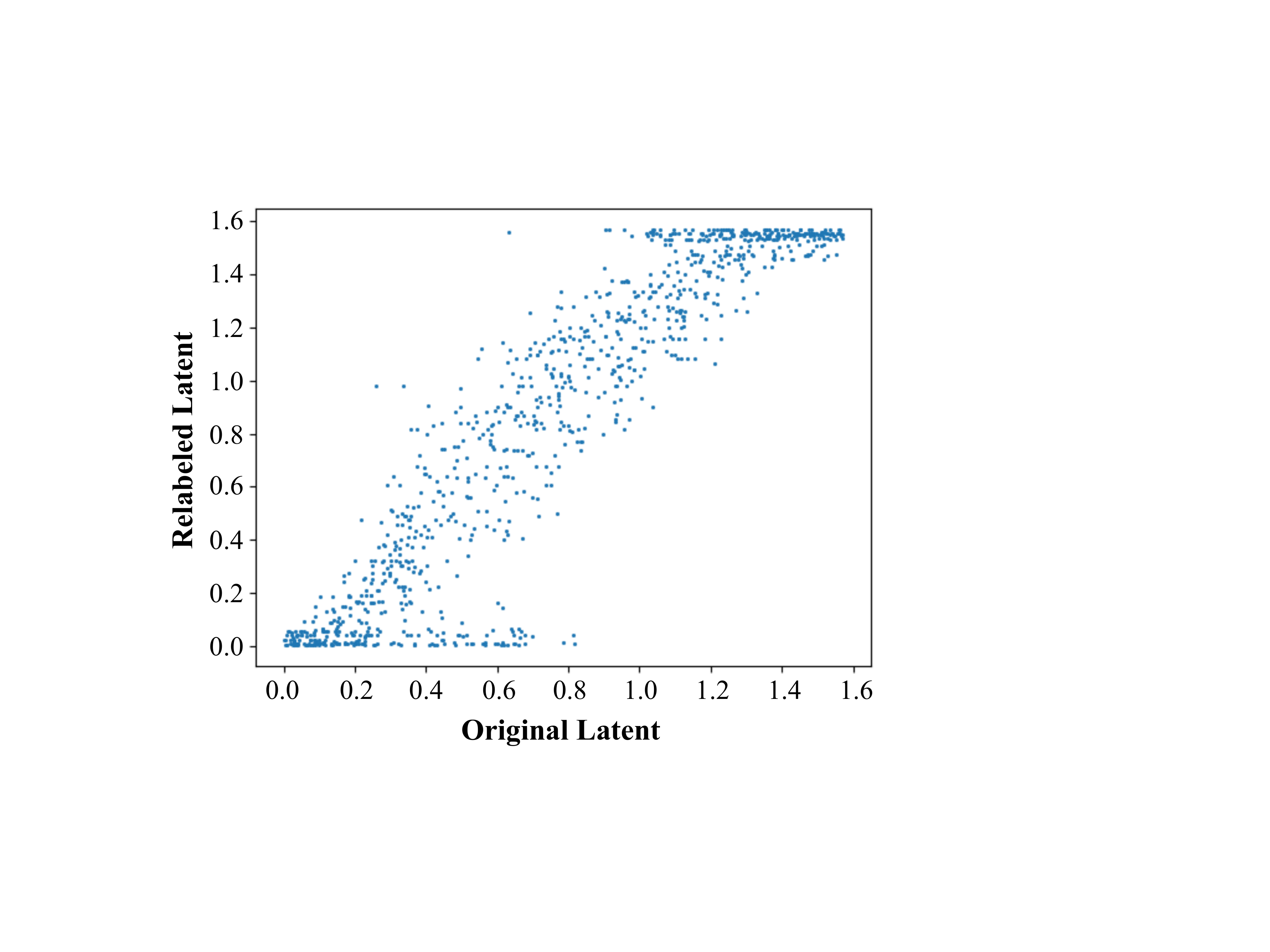}}
\subfigure[Advantage Relabeling]{\includegraphics[width=0.33\linewidth]{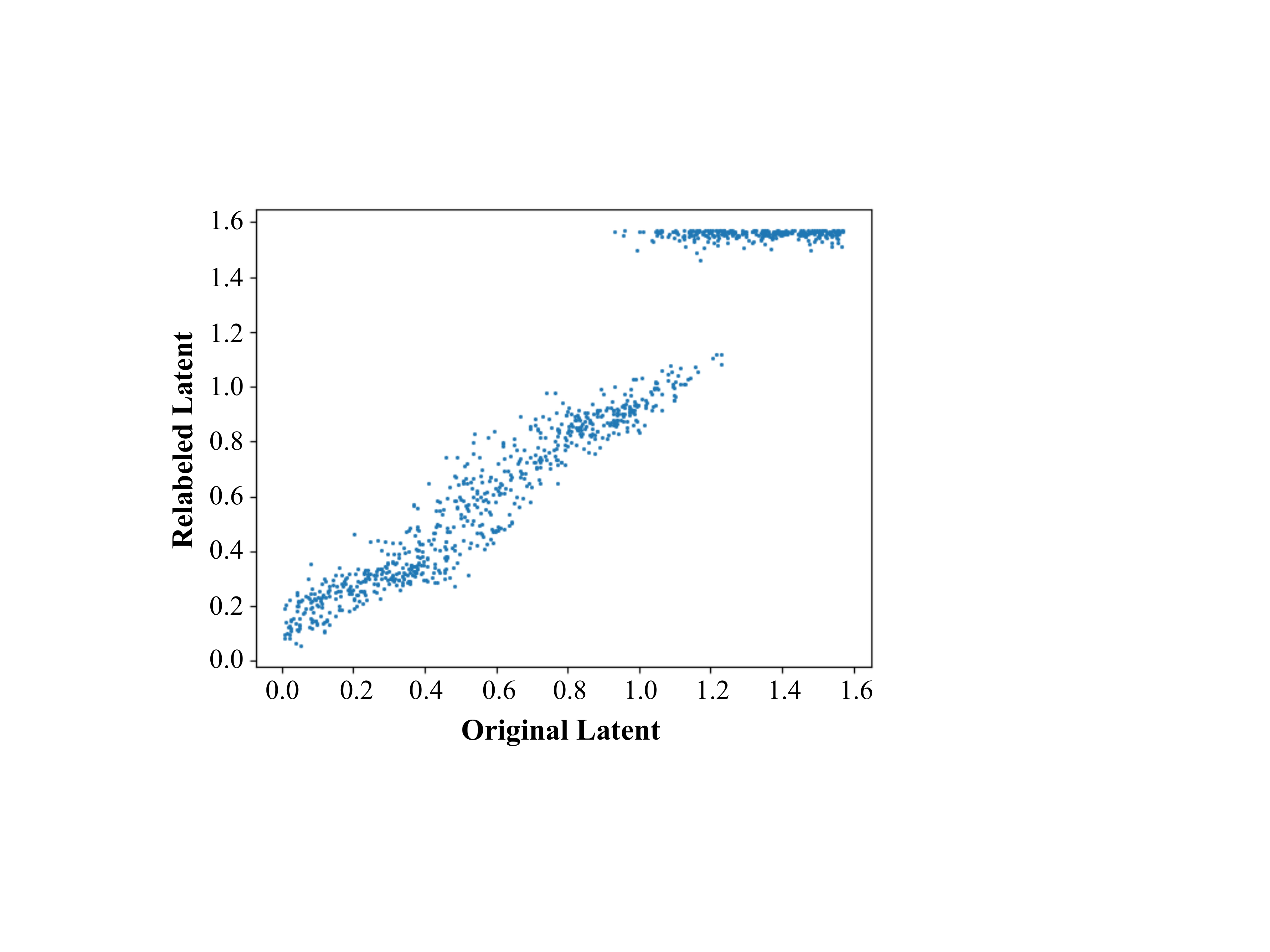}}
\subfigure[Reward Relabeling]{\label{fig:relabeling}\includegraphics[width=0.33\linewidth]{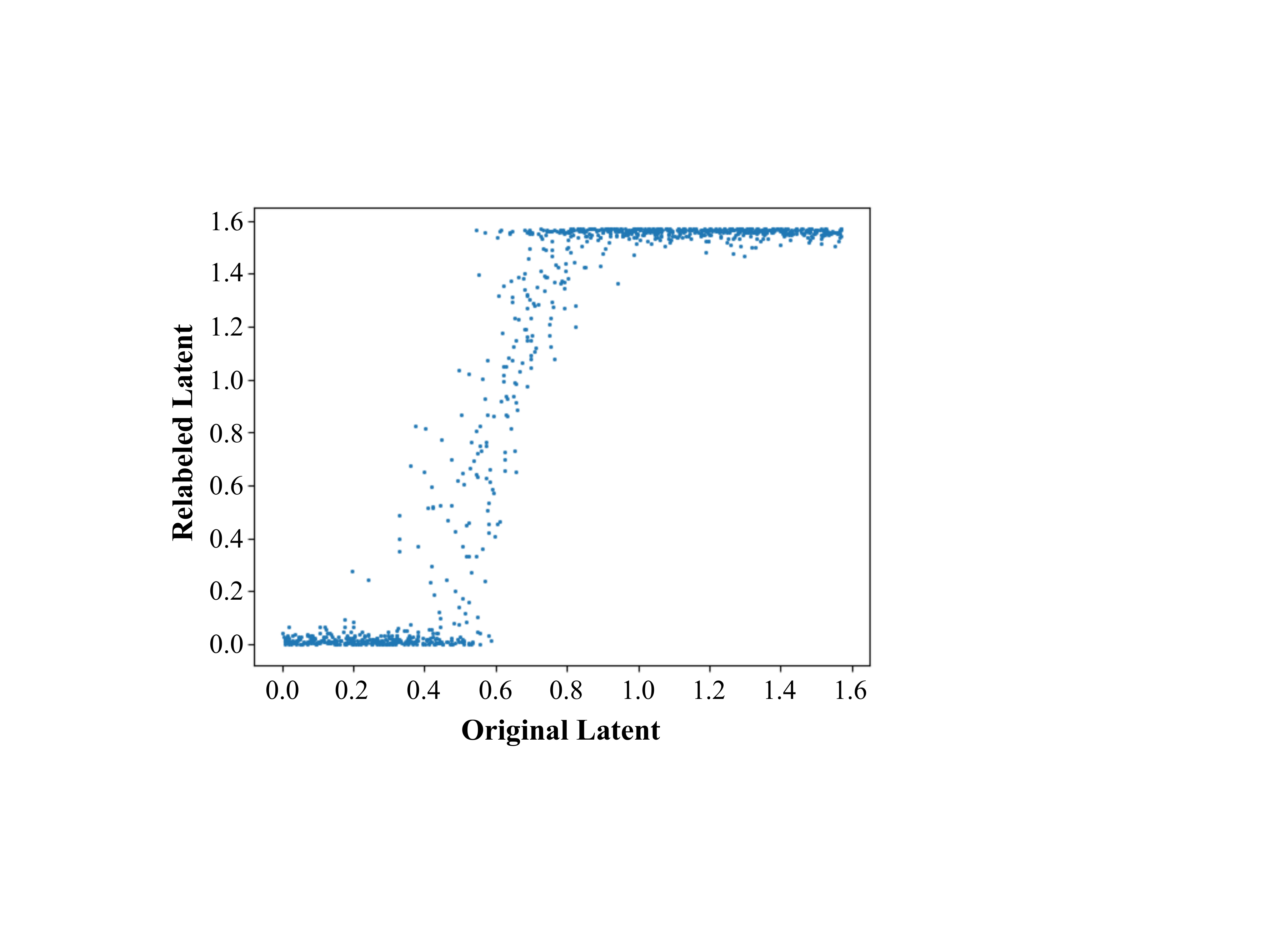}}
\caption{Comparison of relabeling fidelity on optimal trajectories for approximate IRL, advantage relabeling, and reward relabeling. We train a multi-task policy to convergence on the PointReacher environment. We roll out our policy on 1000 randomly sampled tasks $z$, and apply each relabeling method to select from $K=100$ randomly sampled tasks $v$. For approximate IRL, we compare against $N=10$ prior trajectories. The x-axis shows the weight on energy for the task $z$ used for the rollout, while the y-axis shows the weight on energy for the relabeled task $z'$. Note that goal location, obstacle location, and weights on their rewards/penalties are varying as well, but are not shown. Closer to to the line $y=x$ indicates higher fidelity, since it implies $z' \approx z^*$.}
\label{fig:relabelingcomparison}
\end{figure*}

\subsection{Analysis of Relabeling Fidelity}
Approximate IRL, advantage relabeling, and reward relabeling are all approximate methods for finding the optimal task $z^*$ that a trajectory is (close to) optimal for. As a result, an important characteristic is their \textit{fidelity}, i.e. how close the $z'$ they choose is to the true $z^*$. In \autoref{fig:relabelingcomparison}, we compare the fidelities of these three algorithms. Approximate IRL comes fairly close to reproducing the true $z^*$, albeit a bit noisily because it relies on the comparison to $N$ past trajectories. Advantage relabeling is slightly more precise, but fails for large energy weights, likely because the value function is not precise enough to differentiate between these tasks. Finally, reward relabeling does poorly, since it naively assigns $z'$ solely based on the trajectory reward, not how close the trajectory reward is to being optimal.

%% file: related_work.tex
\section{Related Work}

\subsection{Multi-task and transfer learning}
Learning models that can share information across tasks has been concretely studied in the context multi-task learning~\cite{caruana1997multitask}, where models for multiple tasks are simultaneously learned. More recently, \citet{kokkinos2017ubernet, doersch2017multi} looks at shared learning across visual tasks, while \citet{devin2017learning,pinto2017learning} looks at shared learning across robotic tasks. 

Transfer learning~\cite{pan2009survey,torrey2010transfer} focuses on transferring knowledge from one domain to another. One of the simplest forms of transfer is finetuning~\cite{girshick2014rich}, where instead of learning a task from scratch it is initialized on a different task. Several other works look at more complex forms of transfer~\cite{yang2007adapting,hoffman2014continuous,aytar2011tabula,saenko2010adapting,kulis2011you,fernando2013unsupervised,gopalan2011domain,jhuo2012robust}. 

In the context of RL, transfer learning~\cite{taylor2009transfer} research has focused on learning transferable features across tasks~\cite{parisotto2015actor,barreto2017successor,omidshafiei2017deep}.  Another line of work by \cite{rusu2016progressive,kansky2017schema,devin2017learning} has focused on network architectures that improves transfer of RL policies. Another way of getting generalizable policies is through domain randomization~\cite{sadeghi2016cad,tobin2017domain}, i.e. train an unconditional policy across all of the domains in the multi-task learning setting. Although this works for task distributions over the dynamics and observation space~\cite{pinto2017asymmetric}, it cannot handle distributions of reward functions as seen in our experiments. The techniques of domain randomization are however complementary to our method, where it can provide generalizability to dynamics and observation space while Generalized Hindsight can provide generalizability to different reward functions. 

Hierarchical reinforcement learning~\cite{morimoto2001acquisition, barto2004intrinsically} is another framework amenable for multitask learning. Here the key idea is to have a hierarchy of controllers. One such setup is the Options framework~\cite{sutton1999between} where the higher level controllers breaks down a task into sub-tasks and chooses a low-level controller to complete that sub-task. 
Variants of the Options framework \cite{frans2017meta, li2019sub} have examined how to train hierarchies in a multi-task setting, but information re-use across tasks remains restricted to learning transferable primitives. Generalized Hindsight could be used to train these hierarchical policies more efficiently. 

\subsection{Hindsight in RL}
Hindsight methods have been used for improving learning across as variety of applications. \citet{andrychowicz2017hindsight} uses hindsight to efficiently learn on sparse, goal-conditioned tasks. \citet{nair2018visual} approaches goal-reaching with visual input by learning a latent space encoding for images, and using hindsight relabeling within that latent space. \citet{ding2019goal} uses hindsight relabeling as a form of data augmentation for a limited set of expert trajectories, which it then uses with adversarial imitation learning to learn goal-conditioned policies. Several hierarchical methods \cite{levy2017hierarchical, nachum2018data} train a low-level policy to achieve subgoals and a higher-level controller to propose those subgoals. These methods use hindsight relabeling to help the higher-level learn, even when the low-level policy fails to achieve the desired subgoals. Generalized Hindsight could be used to allow for richer low-level reward functions, potentially allowing for more expressive hierarchical policies. 

\subsection{Inverse Reinforcement Learning}
Inverse reinforcement learning (IRL) has had a rich history of solving challenging robotics problems~\cite{abbeel2004apprenticeship,ng2000algorithms}. More recently, powerful function approximators have enabled more general purpose IRL. For instance, \citet{ho2016generative} use an adversarial framework to approximate the reward function. \citet{li2017infogail} build on top of this idea by learning reward functions on demonstrations from a mixture of experts. Although, our relabeling strategies currently build on top of max-margin based IRL~\cite{ratliff2006maximum}, our central idea is orthogonal to the choice of IRL techniques and can be combined with more complex function approximators.

%% file: conclusion.tex
\section{Conclusion}

In this work, we have presented Generalized Hindsight, an approximate IRL based task relabelling algorithm for multi-task RL. We demonstrate how efficient relabeling strategies can significantly improve performance on simulated navigation and manipulation tasks. Through these first steps, we believe that this technique can be extended to other domains like real world robotics, where a balance between different specifications, such as energy use or safety, is important. 

\textbf{Acknowledgements} We thank AWS for computing resources. We also gratefully
acknowledge the support from Berkeley DeepDrive, NSF, and the ONR Pecase award.

%% file: appendix.tex
\begin{appendices}

\section{Environment Descriptions}
\textbf{PointTrajectory}: 
\begin{itemize}
    \item Dynamics: This environment requires controlling a pointmass on a 2D plane, where the state $s_t = (x_t,y_t)$ represents the location of the pointmass. It always starts at the origin, with $s_0 = (0,0)$, and the agent is restricted to the box $[-1, 1]^2$ via clipping. The agent can take actions $(dx, dy) \in [-0.1, 0.1]^2$, which affect the state via $s_{t+1} = (x_t + dx, y_t + dy)$.
    \item Rewards: The agent is rewarded for following a sinusoidal trajectory parameterized by a three-dimensional vector $z = (\theta, d, a)$. $\theta$ controls the orientation of the sinusoidal trajectory, $d$ controls its wavelength, and $a$ controls its amplitude. Specifically, the reward function is:
    \begin{align*}
        r(s, a | z) = \begin{cases}
        \frac{\tilde x}{d} \phi(\frac {\tilde y - a \times sin(\pi \tilde x / d)}{0.05}) & \text{if } \tilde{x } \geq 0 \\
        0 & \text{if } \tilde{x } < 0
        \end{cases}
    \end{align*}
    where $\tilde x$ is the projection of the current state $(x,y)$ onto the line $y = \tan (\theta) x$, $\tilde y$ is the orthogonal component, and $\phi$ is the probability density function of the unit Gaussian. The $x/d$ term encourages movement towards the goal, and the $\phi(\cdot)$ term sharply penalizes an agent for deviating from the target trajectory.
    
    The task distribution $\mathcal T$ is as follows: each element of $z$ is drawn independently from a separate distribution, where $\theta \sim \text{Unif}[-\pi, \pi]$, $d \sim \text{Unif}[0.75, 1]$, and $a \sim \text{Unif}[-0.25, 0.25]$.
\end{itemize}

\textbf{PointReacher}
\begin{itemize}
    \item Dynamics: PointReacher shares the same dynamics as PointTrajectory.
    \item Rewards: This environment requires managing three quantities of interest: distance to a goal, distance from an obstacle, and energy used. We control these via a 6-dimensional task vector $z= (x_g,y_g,x_{obst},y_{obst},u,v)$, where the goal is located at $(x_g, y_g)$, the obstacle is located at $(x_{obst}, y_{obst})$, and $u$ and $v$ control the relative weighting of the three terms. Specifically, $u$ and $v$ represent a location on a unit sphere, and we calculate the weights $w_1$, $w_2$, and $w_3$ as the Euclidean coordinates of that point. The distance reward $r_{goal, t}$, safety reward $r_{obst, t}$, and energy penalty $E_t$ are calculated as following:
    
    \begin{align*}
        r_{goal, t} &= 2 \exp \left\{\frac{-((x_t - x_g)^2 + (y_t - y_g)^2)}{0.08^2} \right\} \\
        r_{obst, t} &= \log_{10}\left(0.01 + (x_t - x_{obst})^2 + (y_t - y_{obst})^2\right) \\
        E_t &= - ||a_t||_2
    \end{align*}
    We found that using these formulations encouraged specificity for the goal (i.e. optimal behavior is to move to the goal, not just a relatively close location), and a heavy penalty for coming too close to the obstacle. 
    
    Overall, the reward is then:
\begin{align*}
    r(s_t, a_t| z) = w_1 r_{goal,t} + w_2 E_t + w_3 r_{obst,t}
\end{align*}
    Note that the obstacle is not physically present in the environment, so the dynamics of the environment are the same across all possible tasks. The obstacle is solely a part of the reward function, and the agent is discouraged from coming nearby via the penalty in the reward function. In general, this represents the idea that we can ``practice'' for certain safety-critical applications without needing to actually interact with the dangerous obstacle at hand. 
    
    The task distribution $\mathcal T$ is as follows: $u,v$ are sampled uniformly from the portion of the sphere in the first octant, and $(x_g, y_g)$ and $(x_{obst}, y_{obst})$ are sampled uniformly from the disk of radius 0.3 centered at the start state $s_0 = (0,0)$.
\end{itemize}

\textbf{Fetch}
\begin{itemize}
    \item Dynamics: We use the Fetch Robot from OpenAI Gym \cite{brockman2016openai, plappert2018multi}. Fetch has 3-dimensional observations, corresponding to the coordinates of its end-effector. It is operated through noisy position control, so the action space is also 3-dimensional. We increase the number of solver iterations from 20 to 100, to make the controller more accurate and reduce control noise.
    
    \item Rewards: We use the same parameterized reward function as we do in PointReacher, adapted so that the goal and obstacle lie in 3D space. 
\end{itemize}

\textbf{HalfCheetahMultiObjective}
\begin{itemize}
    \item Dynamics: We use the HalfCheetah-v1 environment from OpenAI Gym \cite{brockman2016openai}, corresponding to a two-legged robot constrained to run in the $xz$ plane. It has a 17-dimensional observation space, and a 6-dimensional action space of torque inputs for each joint. 
    \item Rewards: Our task variable $z \in \mathcal Z \subseteq \mathbb{R}^4$ controls a weighted combination of velocity in the x-direction ($v_t$), energy use ($E_t$), height of the center of mass ($h_t$), and rotation speed ($\omega_t)$. 
    \begin{align*}
        r(s_t, a_t| z) = z_1 v_t + z_2 E_t + z_3 h_t + z_4 \omega_t
    \end{align*}
    where $v_t = x_{t+1} - x_{t}$, $\omega_t = \theta_{t+1} - \theta_{t}$, and $E_t = -0.1 \times ||a_t||_2^2$. Since only the ratio between individual elements of $z$ are relevant for eliciting different behavior, we constrain the set of allowed task variables $\mathcal Z = \{z \in \mathbb R ^4 : ||z||_2  = 1, z_1 \geq 0, z_2 \geq 0\}$. We enforce non-negativity of $z_1$ and $z_2$ because it is rather unimpressive behavior to maximize energy use without purpose or minimize the height of the robot's center of mass. To sample from the task distribution $\mathcal T$, we sample uniformly from $\mathcal Z$.
\end{itemize}

\textbf{AntDirection}
\begin{itemize}
    \item Dynamics: We use the Ant-v2 environment from OpenAI Gym \cite{brockman2016openai}, corresponding to a four-legged robot that can move in the $x$, $y$, and $z$ directions. The state space is 111-dimensional, and the joints are controlled via torque control, for an 8-dimensional action space. 
    \item Rewards: We use a 1-dimensional task variable to control which direction the Ant robot should move in. Our reward function is:
    \begin{align*}
        r(s, a |z) = ||\text{velocity}||_2 \times \mathbbm{1}\{\text{velocity angle within 15 degrees of z}\}
    \end{align*}    
    This reward function encourages moving quickly in the direction chosen by the task variable $z$. An alternative reward function that returns the length of the velocity component in the direction of $z$ allows a policy to gain high rewards while going even at a $45^\circ$ angle from the target direction, so we add the indicator function to restrict the space of high-reward behavior. The task distribution $\mathcal T$ is uniform over $\mathcal Z = [- \pi, \pi]$.
\end{itemize}

\section{Training Details}
\subsection{Training Tricks}

Since our task space $\mathcal Z$ is continuous, we cannot use multi-headed networks, which are commonly used when dealing with a discrete set of tasks \cite{yu2019meta}. We thus follow the ubiquitous approach of concatenating $(s || z)$ and feeding that into the policy network $\pi$, and concatenating $(s || a || z)$ and feeding that into the q-network. However, when the observation space is large, such as in Ant (111-dimensional) or HalfCheetah (17-dimensional), it becomes difficult for the network to identify the few dimensions of its input that correspond to $z$ and should thus strongly determine the desired behavior (for $\pi$) or the correct q-value. 

We deal with this by (a) repeating the task variable $z$ a few times, and (b) appending the repeated latent at every hidden layer. (a) increases the salience of $z$, which should improve the ability of the network to do credit assignment and identify $z$ as the causal variable responsible for the difference between tasks. (b) allows the policy and q networks to more easily have differentiated behavior, depending on the current task $z$. Empirically, we find that these two tricks are crucial for getting the baselines (No relabeling, Intentional-Unintentional Agent) to work at all; our Generalized Hindsight methods benefit to a smaller degree. For HalfCheetahMultiObjective and AntDirection, we repeat the task variable 5 times. 

\subsection{Hyperparameters}
We list shared hyperparameters in \autoref{tab:hyperparams}, and environment-specific hyperparameters in \autoref{tab:envhyperparams}.
\begin{table*}[!htbp]
  \begin{center}
    \begin{tabular}{l|l}\toprule
      Parameter & Value\\
      \hline
      Algorithm & Soft Actor Critic \cite{haarnoja2018soft} \\
      Optimizer & Adam \cite{kingma2014adam} \\
      Batch size & 256 \\
      Target smoothing coefficient ($\tau$) & 0.005\\
      Reward scale & Auto-tuned \cite{haarnoja2018softapplications} \\
    \end{tabular}
  \end{center}
\caption{Hyperparameters used for the experiments shown in \autoref{fig:learningcurves}.}
\label{tab:hyperparams}
\end{table*}

\begin{table*}[!htbp]
  \begin{center}
    \begin{tabular}{l|ccccc}\toprule
      Environment & Hidden Layers & Learning Rate & Gradient Steps per Epoch & Horizon  & Discount ($\gamma$)\\
      \hline
      PointTrajectory &     [400, 300] & $3 \times 10^{-3}$ & 100 & 15 & 0.9\\
      PointReacher &        [400, 300] & $3 \times 10^{-3}$ & 200 & 20 & 0.97\\
      Fetch &               [400, 300] & $3 \times 10^{-3}$ & 200 & 50 & 0.98\\
      HalfCheetahMultiObj & [256, 256] & $3 \times 10^{-4}$ & 1000 & 1000 & 0.99\\
      AntDirection &        [256, 256] & $3 \times 10^{-4}$ & 1000 & 1000 & 0.99
    \end{tabular}
  \end{center}
\caption{Environment-specific hyperparameters used for the experiments shown in \autoref{fig:learningcurves}.}
\label{tab:envhyperparams}
\end{table*}

\section{Videos}
A video of our environments and results can be found here: \href{https://sites.google.com/view/generalized-hindsight}{sites.google.com/view/generalized-hindsight}.

\end{appendices}